\documentclass{article} 
\usepackage{graphicx}
\usepackage{multirow}
\usepackage{epic,eepic,color}
\definecolor{red}{rgb}{1,0,0}
\definecolor{green}{rgb}{0,1,0}
\definecolor{blue}{rgb}{0,0,1}
\definecolor{violet}{rgb}{1,0,1}
\definecolor{cyan}{cmyk}{1,0,0,0}
\definecolor{magenta}{cmyk}{0,1,0,0}
\definecolor{yellow}{cmyk}{0,0,1,0}

\definecolor{white}{rgb}{1,1,1}

\onecolumn \textwidth=17cm  

\begin{document}

\newcommand{\FIG}[3]{
\begin{minipage}[b]{#1cm}
\begin{center}
\includegraphics[width=#1cm]{#2}
{\scriptsize #3}
\end{center}
\end{minipage}
}

\newcommand{\FIGm}[3]{
\begin{minipage}[b]{#1cm}
\begin{center}
\includegraphics[width=#1cm]{#2}\vspace*{-2mm}\\
{\scriptsize #3}
\end{center}
\end{minipage}
}

\newcommand{\FIGR}[3]{
\begin{minipage}[b]{#1cm}
\begin{center}
\includegraphics[angle=-90,clip,width=#1cm]{#2}\vspace*{1mm}
\\
{\scriptsize #3}
\vspace*{1mm}
\end{center}
\end{minipage}
}

\newcommand{\FIGRpng}[5]{
\begin{minipage}[b]{#1cm}
\begin{center}
\includegraphics[bb=0 0 #4 #5, angle=-90,clip,width=#1cm]{#2}\vspace*{1mm}
\\
{\scriptsize #3}
\vspace*{1mm}
\end{center}
\end{minipage}
}

\newcommand{\FIGpng}[5]{
\begin{minipage}[b]{#1cm}
\begin{center}
\includegraphics[bb=0 0 #4 #5, clip, width=#1cm]{#2}\vspace*{-1mm}\\
{\scriptsize #3}
\vspace*{1mm}
\end{center}
\end{minipage}
}

\newcommand{\FIGtpng}[5]{
\begin{minipage}[t]{#1cm}
\begin{center}
\includegraphics[bb=0 0 #4 #5, clip,width=#1cm]{#2}\vspace*{1mm}
\\
{\scriptsize #3}
\vspace*{1mm}
\end{center}
\end{minipage}
}

\newcommand{\FIGRt}[3]{
\begin{minipage}[t]{#1cm}
\begin{center}
\includegraphics[angle=-90,clip,width=#1cm]{#2}\vspace*{1mm}
\\
{\scriptsize #3}
\vspace*{1mm}
\end{center}
\end{minipage}
}

\newcommand{\FIGRm}[3]{
\begin{minipage}[b]{#1cm}
\begin{center}
\includegraphics[angle=-90,clip,width=#1cm]{#2}\vspace*{0mm}
\\
{\scriptsize #3}
\vspace*{1mm}
\end{center}
\end{minipage}
}

\newcommand{\FIGC}[5]{
\begin{minipage}[b]{#1cm}
\begin{center}
\includegraphics[width=#2cm,height=#3cm]{#4}~$\Longrightarrow$\vspace*{0mm}
\\
{\scriptsize #5}
\vspace*{8mm}
\end{center}
\end{minipage}
}

\newcommand{\FIGf}[3]{
\begin{minipage}[b]{#1cm}
\begin{center}
\fbox{\includegraphics[width=#1cm]{#2}}\vspace*{0.5mm}\\
{\scriptsize #3}
\end{center}
\end{minipage}
}

\title{Leveraging Image based Prior for Visual Place Recognition}

\author{
  Tsukamoto Taisho ~~~~~~ Tanaka Kanji\\
  University of Fukui\\
  3-9-1, Bunkyo, Fukui, Fukui, JAPAN\\
  {\tt tnkknj@u-fukui.ac.jp}\\
}

\maketitle

\newcommand{\figA}{
\begin{figure}[t]
\begin{center}
\begin{center}
\FIGR{8}{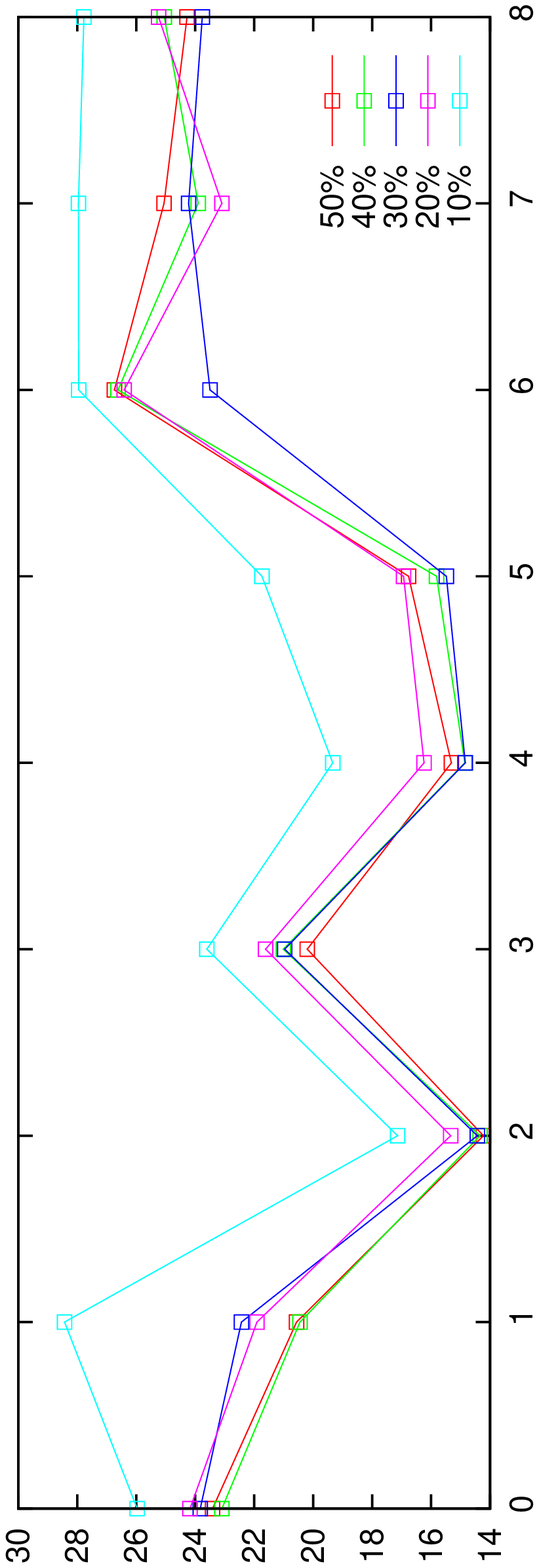}{}
\caption{Graph showing the effect of 
the number of landmarks used per image
during the description process.}\label{fig:A}~~\vspace*{5mm}\\
\FIGR{8}{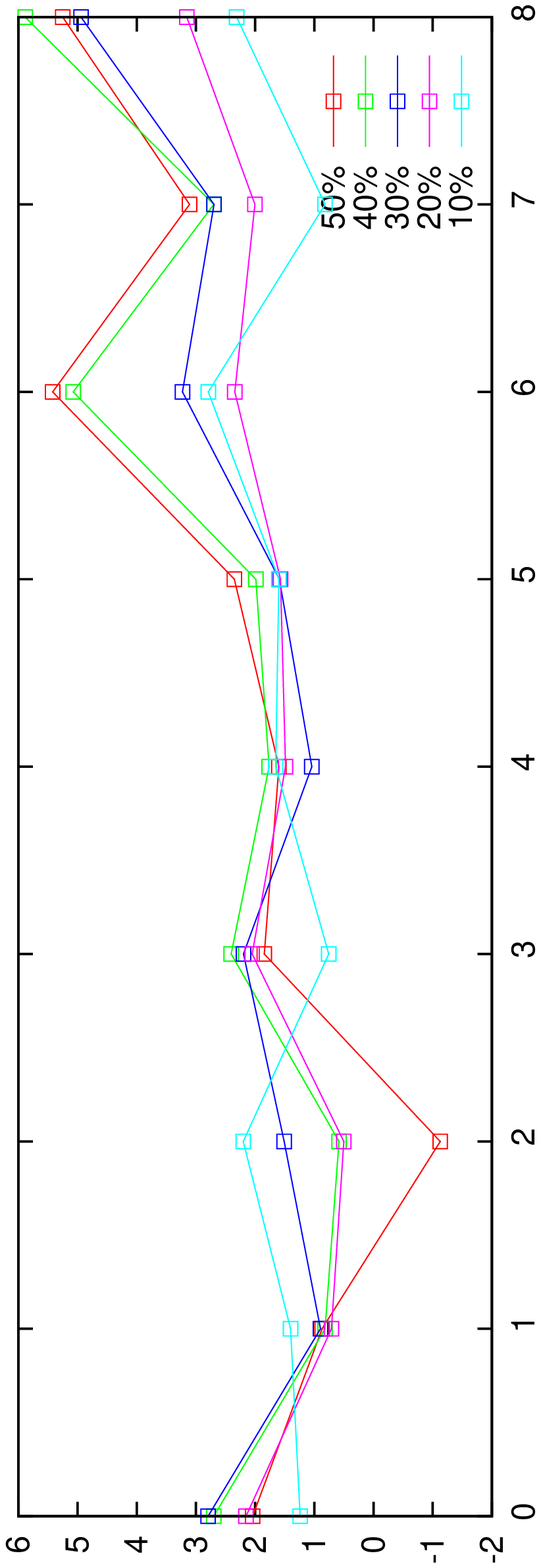}{}
\caption{Comparison between cases with and without BBs.}\label{fig:Ad}
\end{center}
\end{center}
\end{figure}
}

\newcommand{\figB}{
\begin{figure}[t]
\begin{center}
\begin{center}
\FIG{8}{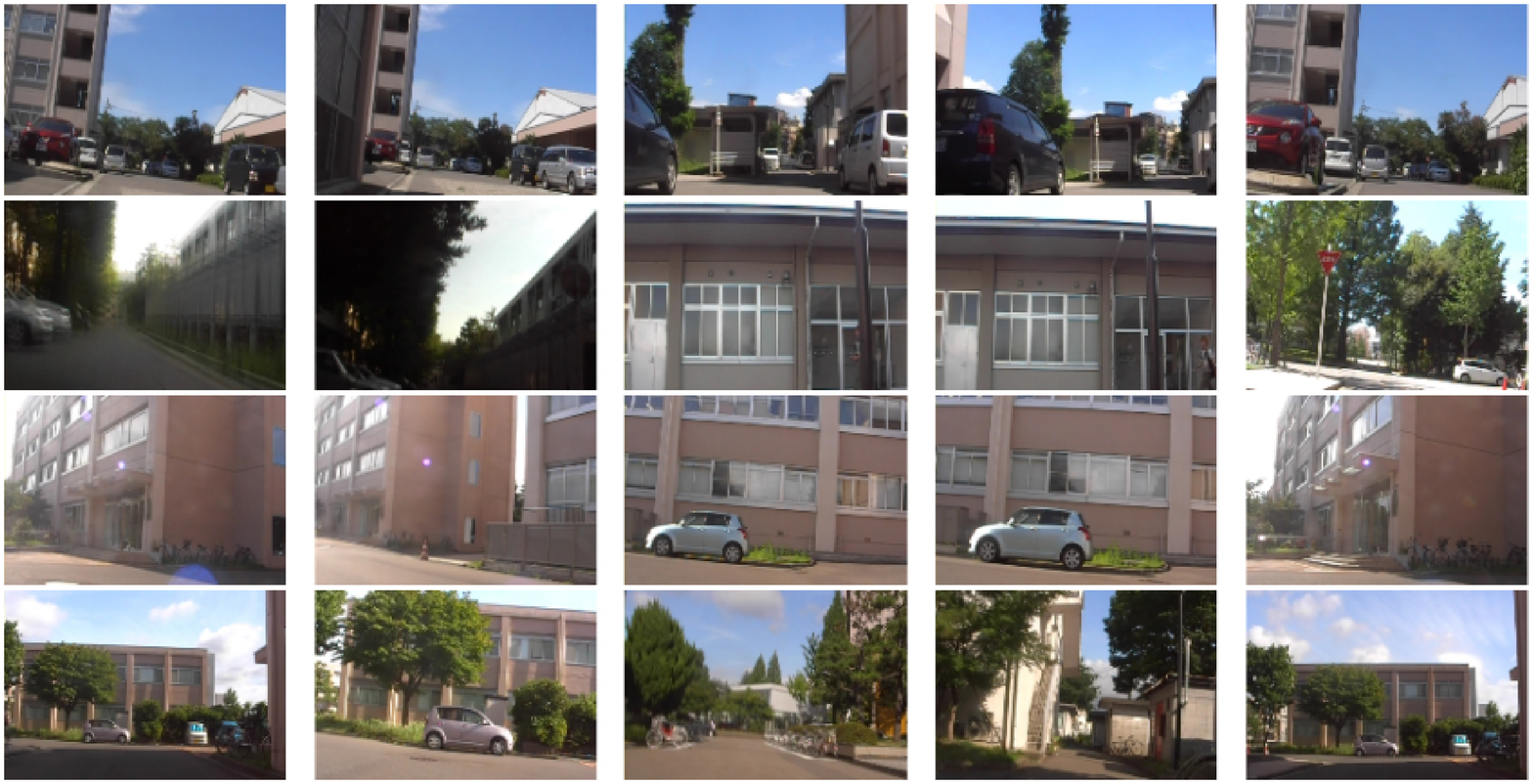}{}
\caption{Snapshot that shows the cases in which our method fails.}\label{fig:B}
\end{center}
\end{center}
\end{figure}
}

\newcommand{\tabCold}{ 
\begin{table}[t]
\begin{center}
{\small
\caption{Performance results.}\label{tab:C} 
\vspace*{3mm}
\begin{tabular}{|c|c|c|c|c|c|c|cl}
\hline
\multirow{2}{*}{\hspace*{-1mm}dataset\hspace*{-1mm}} & \multirow{2}{*}{\hspace*{-1mm}BoW}\hspace*{-1mm}& \multirow{2}{*}{\hspace*{-1mm}VLAD\hspace*{-1mm}} & \multicolumn{2}{ |c| }{IP} \\ \cline{4-5}
& & & w/o BB & w/ BB \\ \hline \hline
 0 & 31.7 & 26.9 & 24.2 & 22.0 \\ \cline{1-5}
 1 & 38.7 & 27.8 & 21.9 & 21.2 \\ \cline{1-5}
 2 & 34.4 & 14.0 & 15.3 & 14.8 \\ \cline{1-5}
 3 & 27.5 & 20.8 & 21.6 & 19.6 \\ \cline{1-5}
 4 & 28.9 & 17.5 & 16.2 & 14.8 \\ \cline{1-5}
 5 & 21.6 & 17.6 & 16.9 & 15.4 \\ \cline{1-5}
 6 & 21.7 & 27.1 & 26.4 & 24.1 \\ \cline{1-5}
 7 & 28.9 & 28.2 & 23.1 & 21.1 \\ \cline{1-5}
 8 & 26.4 & 23.7 & 25.2 & 22.1  \\ \cline{1-5}
\end{tabular}
}
\end{center}
\end{table}
}

\newcommand{\tabC}{ 
\begin{table}[t]
\begin{center}
{\small
\caption{Performance results.}\label{tab:C} 
\vspace*{3mm}
\begin{tabular}{|c|c|c|c|c|c|c|cl}
\hline
\multirow{2}{*}{\hspace*{-1mm}dataset\hspace*{-1mm}} & \multirow{2}{*}{\hspace*{-1mm}BoW}\hspace*{-1mm}& \multirow{2}{*}{\hspace*{-1mm}VLAD\hspace*{-1mm}} & \multicolumn{2}{ |c| }{IP} \\ \cline{4-5}
& & & w/o BB & w/ BB \\ \hline \hline
 0 & 31.7 & 26.9 & 24.2 & 22.0 \\ \cline{1-5}
 1 & 38.7 & 27.8 & 21.9 & 21.2 \\ \cline{1-5}
 2 & 34.4 & 14.0 & 15.3 & 14.8 \\ \cline{1-5}
 3 & 27.5 & 20.8 & 21.6 & 19.6 \\ \cline{1-5}
 4 & 28.9 & 17.5 & 16.2 & 14.8 \\ \cline{1-5}
 5 & 21.6 & 17.6 & 16.9 & 15.4 \\ \cline{1-5}
 6 & 21.7 & 27.1 & 26.4 & 24.1 \\ \cline{1-5}
 7 & 28.9 & 28.2 & 23.1 & 21.1 \\ \cline{1-5}
 8 & 26.4 & 23.7 & 25.2 & 22.1  \\ \cline{1-5}
\end{tabular}
}
\end{center}
\end{table}
}

\newcommand{\figE}{
\begin{figure}[t]
\begin{center}
\FIGR{8}{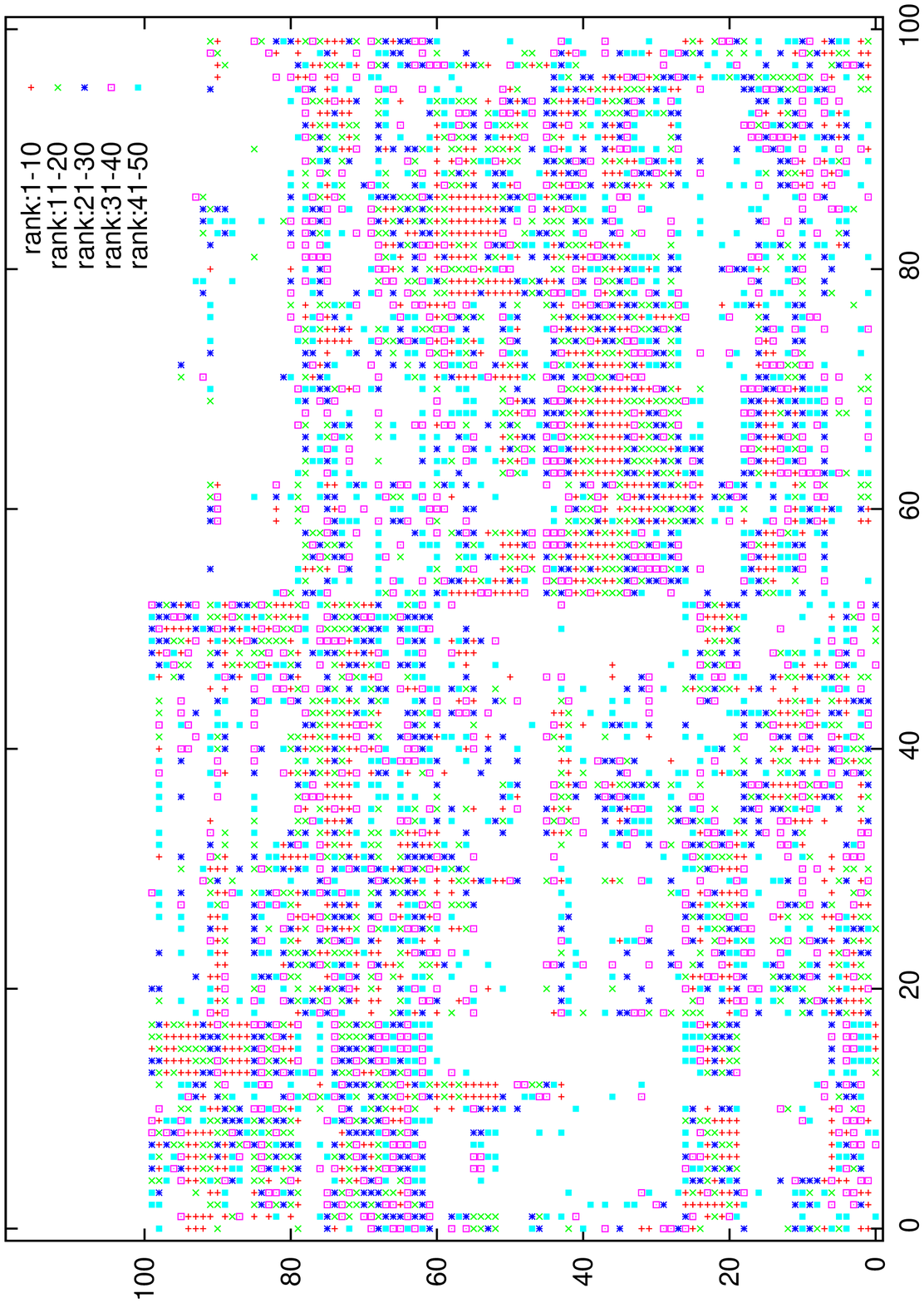}{}
\end{center}
\caption{Relationship between input and library images.
The x-axis represents the ID of input image 
that needs to be classified,
whereas the y-axis indicates the 
ID of library images used.}\label{fig:E}
\end{figure}
}

\newcommand{\x}[1]{\begin{minipage}{6.5mm}
\begin{center}
{\tiny
#1}
\end{center}
\end{minipage}
}

\newcommand{\figF}{
\begin{figure*}[t]
\begin{center}
\FIG{16}{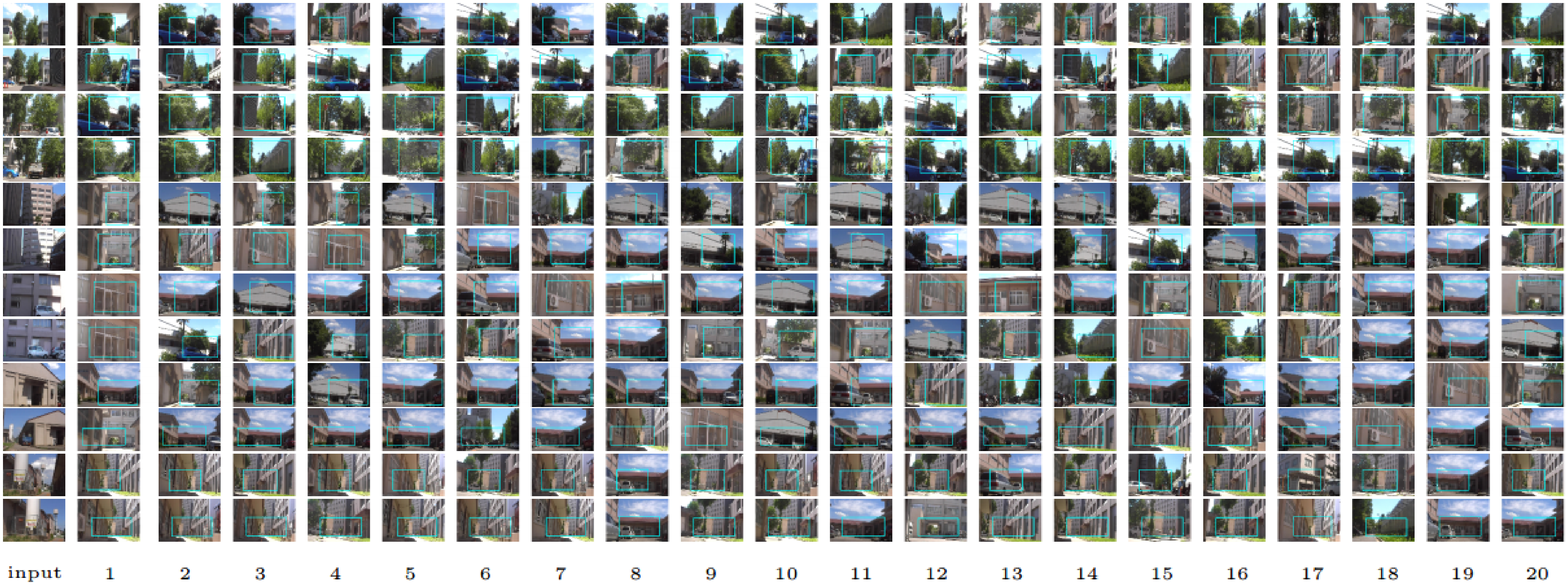}{}
\caption{Scene parsing. The first column shows the input view image 
that is to be described. 
The columns numbered 1-20 show the $L=20$ library images used for describing the input view image. }\label{fig:F}
\end{center}
\end{figure*}
}

\newcommand{\figH}{
\begin{figure}[t]
\begin{center}
\FIG{8}{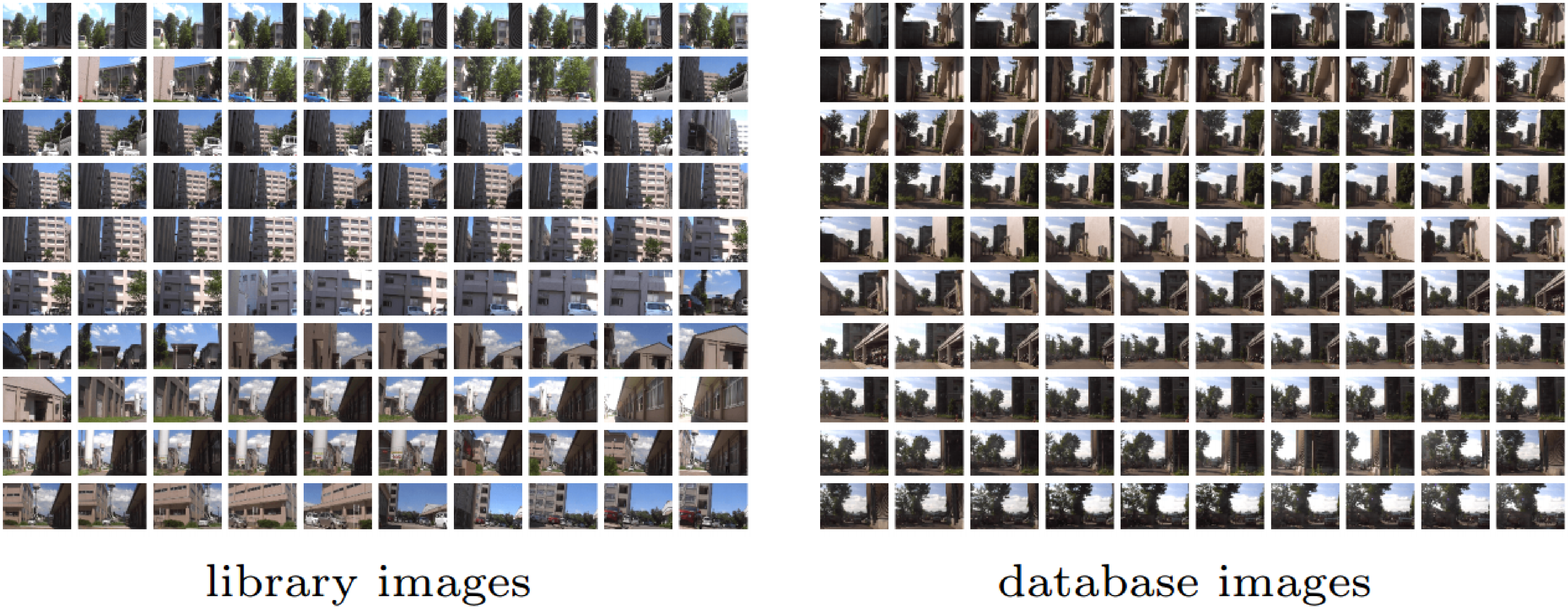}{}
\caption{Snapshot of our image collection captured at a University Campus.}\label{fig:H}
\end{center}
\end{figure}
}

\section*{\centering Abstract}
\textit{
In this study,
we propose a novel scene descriptor 
for visual place recognition.
Unlike popular bag-of-words scene descriptors which rely on a library of vector quantized visual features, 
our proposed descriptor is based on 
a library of raw image data,
such as publicly available photo collections from Google StreetView and Flickr.
The library images need 
not to be associated with spatial information 
regarding the viewpoint and orientation of the scene.
As a result, these images are cheaper than the database images;
in addition, they are readily available.
Our proposed descriptor directly mines the image library 
to discover landmarks (i.e., image patches) 
that suitably match an input query/database image. 
The discovered landmarks are then compactly described by their pose and shape (i.e., library image ID, bounding boxes) 
and used as a compact discriminative scene descriptor for the input image. 
We evaluate the effectiveness of our scene description framework by comparing its performance to that of previous approaches.
}

\section{Introduction}

Scene description is an important first stage in visual place recognition (VPR),
which allows one to search through a pre-built image database
to find visually similar views.
The most popular 
scene description method is to translate each image 
into a bag of vector-quantized visual features, 
termed as visual words, and then apply document retrieval techniques 
that are based on the bag-of-words document model (BoW) \cite{fabmap09}. 
Many recent VPR systems are based on the BoW scene description scheme.
Despite its computational efficiency and robustness, 
these BoW scene descriptor -based VPR systems suffer 
from vector quantization errors,
and often fail to handle the appearance 
changes across views that appear in practice \cite{finegrained}. 

In this study, we address this issue by leveraging image based prior.
Unlike popular BoW scene descriptors which rely on a library of vector quantized visual features, 
our proposed method is based on 
a library of raw image data,
such as publicly available photo collections from Google StreetView and Flickr.
These library images need not be associated with spatial information 
such as the viewpoint and orientation of the scene,
and are thus cheaper than the database images;
furthermore, these library images are readily available,
which is an added advantage.
In our approach,
the descriptor directly mines the image library 
to identify landmarks (i.e., image patches) that 
suitably match an input query/database image. 
The discovered landmarks are then compactly described by their pose and shape, 
i.e., library image ID, bounding boxes (BB), 
and used as a compact discriminative scene descriptor for the input image. 
We evaluate the effectiveness of our scene description framework by comparing its performance to that of previous approaches.

The problem associated with conventional scene descriptors for VPR 
have been studied extensively. 
Local feature approaches such as BoW scene descriptors 
have been widely studied considering various aspects,
including
self-similarity of images \cite{bow1},
quantization errors \cite{bow2},
query expansion \cite{bow3},
database augmentation \cite{bow5},
vocabulary tree \cite{bow4},
global spatial geometric verification as post-processing \cite{philbin2007object},
and pyramid matching to capture spatial context \cite{spm}.
Previous researches on VPR 
have shown 
that the BoW scene model is not sufficiently discriminative 
and is often unsuccessful at capturing 
the appearance changes across views \cite{finegrained}. 
Global feature approaches such as GIST feature descriptor \cite{globaldescriptor1} (in which a scene is represented by a single global feature vector) 
focus on the 
compactness of scene description
and have high matching speeds. 
Other possible representations include those that describe a scene as a collection of meaningful parts, such as object models \cite{object_bank} and part models \cite{shout}.
Although these approaches may potentially provide rich 
information about a scene, existing techniques rely on a large amount of training examples to learn about the models under supervision.
Note that our use of 
a publicly available photo collection (e.g., Flickr) 
is different from 
that of large-scale geo-localization \cite{geoloc}
where the collection
is directly utilized as the database 
rather than a library.

This study is motivated by the authors' previuos works
on a novel data mining approach to scene description \cite{acpr2013ando,vem16,csvpr}.
\cite{acpr2013ando}
built a prototype method called ``common landmark discovery",
in which 
landmark objects are mined through common pattern discovery (CPD) between an input image and known reference images. 
This framework has been further extended for large-scale visual place recognition by introducing efficient CPD techniques in \cite{vem16}.
The data mining approach has been utilized for 
single-view cross-season place recognition in \cite{csvpr},
where objects whose
appearance remain the same across seasons 
are utilized as valid landmarks.
The effectiveness of the scene description framework 
was evaluated 
by comparing its performance to that of previous BoW approaches, 
and by adapting the Naive Bayes Nearest neighbor (NBNN) distance metric \cite{nbnn_da}
to our scene description framework, (``NBNN scene descriptor").
In contrast, 
the current study 
further investigates 
the effectiveness of the proposed approach
from a novel perspective of landmark mining.

\section{VPR Framework}\label{sec:approach}

The VPR framework consists of three main steps, including
scene parsing,  scene description, and scene retrieval.
First, during scene parsing,
an input scene is analyzed,
and landmarks are discovered 
that effectively explain the input image.
Second, the framework describes the input scene
using IDs of the library images 
and BBs that crop landmark objects within each library image.
Scene descriptors are also computed for all images in the image database.
Finally, the third step involves the retrieval of database images
using the computed scene descriptors as the query.

For the above mentioned method, we assume a dictionary or library 
of random $L_o$ view images to be given.
The library images need not required 
associated with spatial information 
such as the viewpoint and orientation.
A small subset of $L (L\le L_o)$ appropriate library images
that are most similar to a given input image
are selected and used to interpret the image.
Our experimental results suggest
that high recognition performance 
tends to be associated with 
the coverage of the database images
provided by these library images.

\subsection{Scene Parsing}

We consider scene parsing as data mining over the image library. 
Our scheme begins by over-segmenting the input scene image into 
a set of $R$ superpixels and clustering them into a set of $K$ scene parts, which will serve as landmark candidates. 
Then, it evaluates the usefulness 
of each landmark region in terms of the saliency of the region. 
It selects $K$ landmark regions with 
the highest usefulness score, 
and translates each of these into a compact VLAD code. 
VLAD codes are also computed for each landmark for all images in the image library. 
Then, the image library is searched using the $K$ VLAD codes as query, 
and a score is assigned to each library image 
in terms of the sum of the reverse rank $\sum_{i=1}^K 1/r_i$ 
of the individual VLAD-based ranking results 
$r_i$$(i=1,\cdots,K)$. For image segmentation, 
$R=72$ superpixels are produced by SLIC superpixel \cite{slic},
and clustered into  
$2R-1$ landmark regions using hierarchical region clustering 
method provided in \cite{vem7}.
For saliency evaluation,
the PCA-based distinctiveness score 
that has been described in \cite{vem8} 
is evaluated for all the SIFTs belonging to the region
and these are summed up to obtain the region's saliency.
To calculate the VLAD codes, method used in \cite{vlad} is employed. 
The number of landmarks $K$ per image controls the reliability-efficiency tradeoff of our data mining and currently was set to a relatively high value $K=40$ (i.e., put weights on reliability)
during our study.

\subsection{Scene Description}

We describe a scene using $L$ landmarks and each landmark 
is described as a pairing of a landmark image ID 
and a BB of landmark region 
with respect to the landmark image. 
The procedure for discovering landmark images 
was as discussed in the previous subsection. 
However, the problem of determining the BB has not been addressed yet. 
In the proposed method, 
we extract sets of SIFT features from the input and the library images, $F_Q$ and $F_L$, 
in addition, the nearest point to each $f\in F_Q$ among 
the $F_L$ points in the 128-dim SIFT descriptor space, 
and then use keypoints $\{(x,y)\}$ of the nearest point 
to compute the BB. 
For noise reduction, only the middle 80\% $x$ (or $y$) values are used for the computation after all the $x$ (or $y$) values are sorted numerically. 
As a consequence, our scene descriptor is of the form:
\begin{equation}
\left\{
\langle
I_i, 
B_i
\rangle
\right\}_{i=1}^L,
\end{equation}
where
$I_i$ is the ID of landmark image,
$B_i$ is the BB
consisting of the top left 
and the bottom right node,
$(x^{min}_i, y^{min}_i)$ 
and $(x^{max}_i, y^{max}_i)$,
of BB.

\subsection{Scene Retrieval}

In this final step, we search the image database and score each database image using the scene descriptor. 
To build the database, the image ID 
$I_i$
with the BB 
$B_i$
for each database image is stored in an inverted file using the element $I_i$ as index. This structure is an array of $L_o$ inverted lists, one per library image ID. For database retrieval, each $I_i$ of a given query image is used as the index and all the database images assigned to the inverted list associated with this $I_i$ are returned. 
To evaluate the similarity between the query and each of the returned database images, 
we use the number of common $I_i$ between the image pair as the primary similarity measure, 
and the area of overlap between the BB pair as the secondary similarity measure. 

\figH

\figF

\section{Experimental Results}

To evaluate our proposed method, 
we used an image dataset consisting of view images 
captured at a university campus, 
using a handheld camera as the vision sensor. 
Occlusion is severe in the scenes, and people and vehicles are dynamic entities occupying the scene. 
We took nine different paths three times each,
to collect three independent collections of images of each path,
and used each of them for query, library and database image collections.
The size of each query and library imageset was 100.
The sizes of the database imagesets were 338, 406, 474, 529, 371, 340, 354, 397 and 328.
Fig.\ref{fig:H} 
shows 
examples
of library and database images.
It can be seen that the database consists of near duplicate images,
which makes our scene retrieval a challenging task.

Fig.\ref{fig:F}
shows some examples of scene parsing.
The first column in 
Fig.\ref{fig:F}
shows the input image 
and the following $L=20$ columns show
the $L$ landmark images and their BBs
that describe the input image.
Further, it is evident
that not all the selected landmark images look similar 
to the input query image they describe.
Despite this fact,
many of the landmark images actually contribute 
to obtaining discriminative scene descriptors
as we report in the following results.

Fig.\ref{fig:E}
shows the relationship between
input and library images.
In the figure,
``rank" 
means that 
the ranking assigned 
by our library image selection 
at the image description stage.
For instance,
when we set $L=20$,
only ``rank:1-10" and ``rank:11-20" 
images are used for description.
We observe that 
only a small subset of library images 
tend to contribute to the retrieval performance.

\figE

\tabC

\figA

Table \ref{tab:C} lists performance results.
We evaluated the proposed image based prior method (``IP") in terms of the retrieval accuracy and compare it with the BoW method (``BoW") \cite{fabmap09},
and VLAD \cite{vlad}.
For the BoW method, 
we employed 
a visual feature descriptor and a vocabulary 
provided in \cite{fabmap09}.
For VLAD, we employed 
the code used in \cite{vlad}.
A series of independent 100$\times$9 retrievals were conducted 
for each of the 100 random query images of all the 9 different paths.
The retrieval performance was measured in terms of the averaged normalized rank (ANR) as percentage; the ANR is a ranking-based retrieval performance measure wherein a smaller value indicates a better retrieval performance. To evaluate ANR, the rank assigned to the ground-truth relevant image was evaluated for 
each of the 100 independent retrievals, and then the rank was normalized 
on the basis of the database size 
and these ranks were averaged over the 100 retrievals.
From Table \ref{tab:C},
one can observe that our approach outperformed 
both BoW and VLAD in most of the retrievals considered in this study.

We also investigate 
the influence of the parameter $L$, 
i.e., the number of landmarks used for scene modeling.
Fig.\ref{fig:A}
shows the ANR performance 
for different settings of the parameter $L$, 
including $L=$
10, 20, 30, 40 and 50.
As can be seen,
the results are comparable
to each other.
An exception
is the case 
where $L=10$,
where 
the number of landmarks are 
too small to make
our bag of landmarks based representation 
less discriminative.

We also investigated the effect of 
using BBs on the retrieval performance. 
In this study,
we conducted another set of experiments
using the proposed scene descriptor without using the BBs,
as a proof-of-concept,
and compared the recognition performance against that of the proposed descriptor.
Fig.\ref{fig:Ad}
shows the comparison of results of the proposed descriptor 
with and without the BBs.
The vertical axis in this figure
is the ANR performance 
of the case using BBs
subtracted from 
that of the case without using BBs.
It can be seen that 
the ANR performance 
shows an improvement when 
the BBs are used 
for most of the cases considered 
in this study.
A notable exception
is the case where
$L$
is set to a relatively large value,
e.g., 50.
This is due to a large number of landmark images
that naturally include 
dissimilar scenes 
as we already showed in Fig.\ref{fig:E},
and BBs of landmarks
with respect to such dissimilar landmark images
provide less meaningful and less reliable information.
However,
it should be noted that 
even such dissimilar landmark images
do actually improve the scene retrieval performance
as we can see in Fig.\ref{fig:A}.

Fig.\ref{fig:B}
reports some examples of failure cases.
For each row,
the first column shows the query images,
the 2nd, 3rd, and 4th columns show
the images that received 
higher similarity score 
than the ground-truth images
when the proposed method was used,
and the last column shows the ground-truth images.
As can be seen,
the proposed approach can be confused 
if some database images
with locally similar
but globally dissimilar structures
that cannot be captured by ``bag-of-X" scene model
are included.
However,
the issue of the globally dissimilar structure
can be mitigated 
by introducing some extension to the BoX model
such as spatial pyramid matching;
this will form part of our future work.

\figB

\section{Conclusions}

The primary contribution of this paper is the proposal of 
a simple and effective 
approach to VPR.
Unlike popular 
BoW scene descriptors which rely on a library of vector quantized visual features, 
our descriptor is based on 
a library of raw image data, 
such as publicly available photo collections from Google StreetView and Flickr;
our method directly mines the library to discover landmarks (i.e., image patches) that effectively explain an input query/database image. 
The discovered landmarks are then compactly described by their pose and shape (i.e., library image ID, BBs) 
and used as a compact discriminative scene descriptor for the input image. 
Experiments using a challenging dataset validate the effectiveness of the proposed approach.


\begin{thebibliography}{10}

\bibitem{fabmap09}
Mark Cummins and Paul Newman.
\newblock Highly scalable appearance -only slam - fab-map 2.0.
\newblock In {\em Robotics: Science and Systems}, 2009.

\bibitem{finegrained}
Bangpeng Yao, Gary~R. Bradski, and Fei{-}Fei Li.
\newblock A codebook-free and annotation-free approach for fine-grained image
  categorization.
\newblock In {\em CVPR}, pages 3466--3473, 2012.

\bibitem{bow1}
Jan Knopp, Josef Sivic, and Tomas Pajdla.
\newblock Avoiding confusing features in place recognition.
\newblock In {\em ECCV}, pages 748--761. Springer, 2010.

\bibitem{bow2}
Relja Arandjelovic and Andrew Zisserman.
\newblock Three things everyone should know to improve object retrieval.
\newblock In {\em CVPR}, pages 2911--2918, 2012.

\bibitem{bow3}
Ondrej Chum, Andrej Mikulik, Michal Perdoch, and Jiri Matas.
\newblock Total recall ii: Query expansion revisited.
\newblock In {\em CVPR}, pages 889--896, 2011.

\bibitem{bow5}
Panu Turcot and David~G Lowe.
\newblock Better matching with fewer features: The selection of useful features
  in large database recognition problems.
\newblock In {\em ICCV Workshops}, pages 2109--2116, 2009.

\bibitem{bow4}
Grant Schindler, Matthew Brown, and Richard Szeliski.
\newblock City-scale location recognition.
\newblock In {\em CVPR}, pages 1--7, 2007.

\bibitem{philbin2007object}
James Philbin, Ondrej Chum, Michael Isard, Josef Sivic, and Andrew Zisserman.
\newblock Object retrieval with large vocabularies and fast spatial matching.
\newblock In {\em CVPR}, pages 1--8, 2007.

\bibitem{spm}
Svetlana Lazebnik, Cordelia Schmid, Jean Ponce, et~al.
\newblock Spatial pyramid matching.
\newblock {\em Object Categorization: Computer and Human Vision Perspectives},
  3:4, 2009.

\bibitem{globaldescriptor1}
Matthijs Douze, Herv{\'e} J{\'e}gou, Harsimrat Sandhawalia, Laurent Amsaleg,
  and Cordelia Schmid.
\newblock Evaluation of gist descriptors for web-scale image search.
\newblock In {\em Proceedings of the ACM International Conference on Image and
  Video Retrieval}, pages 19:1--19:8, 2009.

\bibitem{object_bank}
Li-Jia Li, Hao Su, Li~Fei-Fei, and Eric~P Xing.
\newblock Object bank: A high-level image representation for scene
  classification \& semantic feature sparsification.
\newblock In {\em Advances in neural information processing systems}, pages
  1378--1386, 2010.

\bibitem{shout}
Mayank Juneja, Andrea Vedaldi, C.~V. Jawahar, and Andrew Zisserman.
\newblock Blocks that shout: Distinctive parts for scene classification.
\newblock In {\em CVPR}, pages 923--930, 2013.

\bibitem{geoloc}
Amir~Roshan Zamir and Mubarak Shah.
\newblock Accurate image localization based on google maps street view.
\newblock In {\em Computer Vision--ECCV 2010}, pages 255--268. Springer, 2010.

\bibitem{acpr2013ando}
Ando Masatoshi, Tanaka Kanji, Inagaki Yousuke, Chokushi Yuuto, and Hanada
  Shogo.
\newblock Common landmark discovery for object-level view image retrieval:
  Modeling and matching of scenes via bag-of-bounding-boxes.
\newblock In {\em ACPR}, 2013.

\bibitem{vem16}
Tanaka Kanji, Chokushi Yuuto, and Ando Masatoshi.
\newblock Mining visual phrases for long-term visual slam.
\newblock In {\em IROS}, pages 136--142, 2014.

\bibitem{csvpr}
Ando Masatoshi, Chokushi Yuuto, Tanaka Kanji, and Yanagihara Kentaro.
\newblock Leveraging image-based prior in cross-season place recognition.
\newblock In {\em ICRA}, 2015.

\bibitem{nbnn_da}
Tatiana Tommasi and Barbara Caputo.
\newblock Frustratingly easy nbnn domain adaptation.
\newblock In {\em ICCV}, pages 897--904, 2013.

\bibitem{slic}
Radhakrishna Achanta, Appu Shaji, Kevin Smith, Aurelien Lucchi, Pascal Fua, and
  Sabine Susstrunk.
\newblock Slic superpixels compared to state-of-the-art superpixel methods.
\newblock {\em IEEE Trans. PAMI}, 34(11):2274--2282, 2012.

\bibitem{vem7}
Koen~EA Van~de Sande, Jasper~RR Uijlings, Theo Gevers, and Arnold~WM Smeulders.
\newblock Segmentation as selective search for object recognition.
\newblock In {\em ICCV}, pages 1879--1886, 2011.

\bibitem{vem8}
Ran Margolin, Ayellet Tal, and Lihi Zelnik-Manor.
\newblock What makes a patch distinct?
\newblock In {\em CVPR}, pages 1139--1146, 2013.

\bibitem{vlad}
Herv{\'e} J{\'e}gou, Florent Perronnin, Matthijs Douze, Jorge S{\'a}nchez,
  Patrick P{\'e}rez, and Cordelia Schmid.
\newblock Aggregating local image descriptors into compact codes.
\newblock {\em IEEE Trans. PAMI}, 34(9):1704--1716, 2012.

\end{thebibliography}

\end{document}